%% file: main.tex
\documentclass[10pt,twocolumn,letterpaper]{article}

\usepackage{cvpr}              

\usepackage{fontawesome5}
\usepackage{xcolor}
\newcommand{\gold}{\textcolor[RGB]{253,202, 41}{\faMedal}}
\newcommand{\silver}{\textcolor[RGB]{226,223,230}{\faMedal}}
\newcommand{\bronze}{\textcolor[RGB]{237,146, 76}{\faMedal}}
\usepackage[table]{xcolor}
\usepackage{tcolorbox}
\input{preamble}
\definecolor{cvprblue}{rgb}{0.21,0.49,0.74}
\usepackage[pagebackref,breaklinks,colorlinks,allcolors=cvprblue]{hyperref}

\def\paperID{4222} 
\def\confName{CVPR}
\def\confYear{2026}

\title{ChemLabs on ChemO: A Multi-Agent System for Multimodal Reasoning on IChO 2025}

\author{
    Qiang Xu$^{*}$ \and 
    Shenyuan Bai$^{*}$ \and 
    Leqing Chen \and 
    Zijing Liu \and 
    Yu Li$^{\dagger}$ \\
    International Digital Economy Academy, Shenzhen, China \\
    $^{\dagger}$Corresponding Author: liyu@idea.edu.cn\\
}

\usepackage{adjustbox}
\usepackage{multirow}
\usepackage{tabularx}
\usepackage{booktabs}
\usepackage{listings}
\usepackage{xcolor}
\usepackage{xspace}
\usepackage{threeparttable}
\tcbuselibrary{listings,breakable}
\usepackage{hyperref}

\newtcolorbox{ichoproblem}[1][]{
  colback=gray!3,
  colframe=gray!50,
  boxrule=0.8pt,
  arc=1pt,
  left=3pt,
  right=3pt,
  top=2pt,
  bottom=2pt,
  fonttitle=\small,
  title=#1,
  breakable
}

\lstdefinelanguage{json}{
  basicstyle=\ttfamily\small,
  showstringspaces=false,
  breaklines=true,
  morestring=[b]",
  morecomment=[l]{//},
  morecomment=[s]{/*}{*/},
  sensitive=true,
  literate=
   *{true}{{{\color{blue}true}}}{4}
    {false}{{{\color{blue}false}}}{5}
    {null}{{{\color{blue}null}}}{4}
}

\newtcblisting{chemjson}{
  listing engine=listings,
  listing only,
  breakable,
  colback=blue!3,
  colframe=blue!30,
  boxrule=2pt,
  arc=0.5pt,
  fonttitle=\small,
  title=\textbf{Structured Visual Enhancement JSON},
  listing options={
    language=json,
    basicstyle=\ttfamily\small,
    numbers=left,
    numberstyle=\tiny,
    numbersep=5pt,
    showstringspaces=false,
    breaklines=true
  }
}

\newcommand{\ours}{ChemLabs\xspace}

\begin{document}
\maketitle
\input{sec/0_abstract}    
\input{sec/1_intro}
\input{sec/2_related}

\input{sec/3_benchmark}
\input{sec/4_agent}
\input{table/table_2}
\input{sec/5_experiments}

\input{sec/7_analysis}
\input{sec/8_conclusion}

{
    \small
    \bibliographystyle{ieeenat_fullname}
    \bibliography{main}
}

\input{sec/X_suppl}

\end{document}

%% file: sec/0_abstract.tex
\begin{abstract}
Olympiad-level benchmarks in mathematics and physics are crucial testbeds for advanced AI reasoning, but chemistry, with its unique multimodal symbolic language, has remained an open challenge. We introduce ChemO, a new benchmark built from the International Chemistry Olympiad (IChO) 2025. ChemO features two key innovations for automated assessment: Assessment-Equivalent Reformulation (AER), which converts problems requiring visual outputs (e.g., drawing molecules) into computationally tractable formats, and Structured Visual Enhancement (SVE), a diagnostic mechanism to disentangle a model's visual perception capabilities from its core chemical reasoning. To tackle this benchmark, we propose ChemLabs, a hierarchical multi-agent framework that mimics human expert collaboration through specialized agents for problem decomposition, perception, reasoning, and auditing. Experiments on state-of-the-art multimodal models demonstrate that combining SVE with our multi-agent system yields dramatic performance gains. Our top configuration achieves a score of 93.6 out of 100, surpassing an estimated human gold medal threshold and establishing a new state-of-the-art in automated chemical problem-solving. 
ChemO Dataset: \url{https://huggingface.co/datasets/IDEA-AI4SCI/ChemO}
\end{abstract}

%% file: sec/1_intro.tex
\section{Introduction}
\label{sec:intro}

Recent advances in Multimodal Large Language Models (MLLMs) have led to agents achieving gold-medal performance in scientific Olympiads for mathematics~\cite{olympiadbench} and physics~\cite{physics_supernova}, demonstrating superhuman capabilities in complex reasoning. However, the field of chemistry presents a unique and arguably more complex frontier that remains largely unsolved. Unlike other domains, chemistry relies on a dense, symbolic visual language where parsing molecular structures, interpreting reaction schemes, and analyzing spectral data are intertwined with textual context and quantitative calculations. Existing benchmarks like ChemBench~\cite{chembench} do not adequately capture this deep, multimodal reasoning challenge, leaving a critical gap in evaluating and advancing expert-level MLLM.

To address this, we introduce \textbf{ChemO}, a new benchmark built from the International Chemistry Olympiad (IChO) 2025. A core challenge in creating ChemO was handling problems requiring visual outputs (e.g., drawing a molecule). Our solution is \textit{Assessment-Equivalent Reformulation (AER)}, a principled methodology that transforms such problems into formats where models can generate machine-readable outputs (\textit{e.g.}, SMILES strings) rather than visual drawings, while preserving the original assessment criteria. This reformulation enables models to leverage their text generation capabilities to solve problems that would otherwise require visual output modalities. Furthermore, to diagnose model weaknesses, ChemO introduces an optional \textit{Structured Visual Enhancement (SVE)} setting, which provides models with structured textual descriptions of visual elements in the problem (\textit{e.g.}, molecular structures, diagrams), helping to isolate performance bottlenecks between visual perception and core chemical reasoning.

Inspired by how teams of human experts collaborate, we propose \textbf{ChemLabs}, a hierarchical multi-agent system designed to master the ChemO benchmark. As illustrated in Figure~\ref{fig:overview}, ChemLabs decomposes monolithic reasoning into a structured workflow with iterative refinement: a central \texttt{Manager Agent} decomposes problems and dispatches sub-tasks to specialized modules, including a \texttt{Perception Lab} to interpret chemical diagrams, a \texttt{Solving Lab} with domain-specific reasoners, and a two-stage \texttt{Audit Lab} to verify the chemical and logical integrity of proposed solutions. Through multi-agent deliberation and self-correction loops, ChemLabs iteratively refines solutions until they meet rigorous verification criteria. The main contributions of this work are:
\begin{itemize}
    \item We present \textit{ChemO}, the first benchmark to formalize IChO problems. We introduce \textit{AER}, a methodology that reformulates visual output problems into machine-readable formats to enable text-generation-based solving, and \textit{SVE}, which provides structured textual descriptions of visual elements to isolate visual perception from chemical reasoning capabilities.
    \item We propose \textit{ChemLabs}, a hierarchical multi-agent framework that implements a divide-and-conquer strategy with iterative refinement and self-correction mechanisms for complex, multimodal scientific reasoning.
    \item We demonstrate through extensive experiments that ChemLabs, particularly when augmented with SVE and powered by Gemini-2.5 Pro, achieves a score of \textbf{93.6/100} on ChemO, surpassing an estimated human gold-medal threshold and setting a new state-of-the-art. Our findings indicate that accurate visual perception is the primary bottleneck for MLLMs, and that SVE combined with ChemLabs successfully bridges this gap to achieve gold-medal-level performance.
\end{itemize}

%% file: sec/2_related.tex
\section{Related Work}
\label{sec:related_work}

\subsection{Multimodal Reasoning Agents}

Recent advances in MLLMs have enabled AI systems to jointly reason over visual and textual inputs. Foundation models such as GPT-4V~\citep{gpt4v}, Gemini 2.5~\cite{gemini2.5}, and Claude 4.5~\citep{claude4.5} have demonstrated strong performance in visual question answering and diagram understanding, while efficient open-source alternatives like MiniCPM-V~\citep{minicpm-v} and Qwen3-VL~\cite{qwen3} now achieve comparable results.
To enable autonomous problem-solving, agent frameworks like ReAct~\cite{react} integrate reasoning with tool use through iterative thought-action-observation cycles. Modern systems such as LangChain~\citep{langchain} and AutoGPT~\cite{AutoGPT} provide modular architectures for composing agents with planning, memory, and tool integration capabilities. Recent work~\cite{cherian2023deep} has focused on multimodal algorithmic reasoning, where agents must deduce solution strategies for vision-and-language puzzles requiring mathematical and logical skills. Domain-specific implementations like ChemCrow~\cite{chemcrow} have shown that augmenting LLMs with expert-designed tools significantly improves performance on scientific reasoning tasks, motivating our approach for chemistry olympiad problem-solving.

\input{table/table_3}

\subsection{Olympiads as Benchmarks}

As standard benchmarks become saturated, Olympiad-level competitions have emerged as rigorous testbeds for evaluating advanced reasoning in LLMs and MLLMs~\citep{mahdavi2025brains, petrov2025proof, khatibi2025eefsuva, shi2024can, zou2025liveoibench, chen2025rimo, zheng2025livecodebench}. In mathematics, recent systems have achieved gold medal performance at IMO 2025, marking significant progress in formal reasoning capabilities~\citep{he2024olympiadbench, li2025sciagent}. For physics, several benchmarks have been proposed to assess multimodal physical reasoning. OlympiadBench~\citep{he2024olympiadbench} introduces bilingual multimodal scientific problems spanning mathematics and physics at competition level. More recently, HiPhO~\citep{yu2025physicsminions} provides 13 contemporary physics Olympiad exams (2024-2025) with professional evaluation using official grading rubrics, enabling direct comparison with human contestants. Agent-based approaches like Physics Supernova demonstrate competitive results, scoring 23.5/30 on IPhO 2025 theory problems and surpassing the median gold medalist performance~\citep{physics_supernova}. While benchmarks for general chemistry knowledge exist, such as ChemBench~\citep{chembench2024}, they do not capture the unique multimodal reasoning challenges of IChO problems, which integrate complex diagrams, spectra, and reaction schemes. These efforts highlight the potential of Olympiads as challenging, uncontaminated benchmarks for measuring scientific problem-solving abilities~\citep{yu2025physicsminions}, a gap we aim to fill.

%% file: table/table_3.tex
\begin{table*}[t]
\vspace{-0.5cm}

\centering
\fontsize{8}{10}\selectfont
\setlength{\tabcolsep}{5pt}
\renewcommand{\arraystretch}{1}
\begin{tabular}{@{}clcclcc@{}}
\toprule
\textbf{ID} & \textbf{Problem Topic} & \textbf{Field} & \textbf{\#Sub} & \textbf{Problem Type} & \textbf{Difficulty} & \textbf{Representation} \\
\midrule
P1 & Sesquiterpene ozonolysis and rearrangements 
   & Organic Chemistry & 9 & SC(5), QI(4) & \cellcolor{red!60}Hard & rxn, str, mech \\
P2 & Rapamycin stereoselective total synthesis 
   & Organic / Biochemistry & 4 & SC(3), QI(1) & \cellcolor{red!60}Hard & rxn, str, mech \\
P3 & Pd(II) lantern self-assembly structures 
   & Inorganic Chemistry & 7 & SC(1), QI(2), TE(4) & \cellcolor{red!60}Hard & str, graph \\
P4 & Tennis ball pressurization kinetics 
   & Physical Chemistry & 4 & QC(2), MR(2) & \cellcolor{red!10}Easy & text, calc \\
P5 & Solar desalination energy calculations 
   & Physical Chemistry & 8 & QC(8) & \cellcolor{red!10}Easy & text, calc \\
P6 & CO$_2$ reduction pathways comparison 
   & Physical Chemistry & 7 & QI(2), QC(4), MR(1) & \cellcolor{red!10}Easy & text, calc, graph \\
P7 & Crude oil GC-MS fragmentation 
   & Analytical Chemistry & 5 & SC(2), QI(1), QC(2) & \cellcolor{red!30}Medium & ms, str \\
P8 & Catalytic converter CO oxidation 
   & Inorganic / Physical & 9 & SC(2), QI(4), TE(1), QC(2) & \cellcolor{red!60}Hard & rxn, ir, str, graph \\
P9 & Thiamine-dependent enzyme pathways 
   & Biochemistry & 6 & SC(4), QI(1), MR(1) & \cellcolor{red!60}Hard & rxn, str, mech \\
\midrule
\textbf{Total} & & & \textbf{59} & \multicolumn{2}{l}{\textbf{SC(17), QI(15), TE(5), QC(18), MR(4)}} & \\
\bottomrule
\end{tabular}
\caption{Overview of the 9 problems in the ChemO benchmark. \textbf{Problem Types}: Structure Construction (SC), Quantitative Calculation (QC), Qualitative Identification (QI), Tabular Enumeration (TE), and Mechanistic Reasoning (MR). \textbf{Representations in problem}: \texttt{rxn} = reaction scheme, \texttt{str} = molecular structure, \texttt{ms} = mass spectrum, \texttt{ir} = IR spectrum, \texttt{calc} = numerical calculation, \texttt{graph} = graphical data, \texttt{mech} = mechanism arrows, \texttt{text} = descriptive text.}
\label{tab:icho_problems}

\vspace{-0.5cm}
\end{table*}

%% file: sec/3_benchmark.tex
\section{The ChemO Benchmark}

ChemO comprises the IChO 2025 theoretical examination, leveraging the latest publicly available theoretical problems to minimize data contamination risks. We validate the absence of data leakage through rigorous testing Appendix~\ref{app:vert}.

\subsection{Raw Data Collection}
The data preparation process involved systematic extraction from the IChO 2025 theoretical examination. We employed Gemini 2.5 Flash~\citep{gemini2.5} for OCR extraction from PDF documents, capturing problem statements (including textual descriptions, chemical formulas, and visual elements such as molecular structures and spectra) and official solutions with grading rubrics. The OCR output was post-processed to correct common recognition errors in chemical notation (e.g., subscripts, superscripts, special symbols). All extracted content underwent rigorous manual verification by chemistry domain experts to ensure accuracy.

\subsection{Problem and Solution Formalization}

\subsubsection{Multi-modal Problem Representation}

Chemistry olympiad problems exhibit a hierarchical structure with multiple sub-problems. Each atomic problem (the smallest scorable unit) is represented as $\mathcal{P} = \{t, V, a, \mathcal{M}\}$, where:
\begin{itemize}
    \item $t \in \mathcal{T}$ represents the textual component, which includes problem statements, numerical data, chemical formulas, and experimental descriptions.
    \item $V = \{v_1, v_2, \ldots, v_n\}$ denotes the set of visual elements, where each $v_i$ represents an image containing molecular structures, reaction schemes, spectra, diagrams, or other chemical representations.
    \item $a \in \mathbb{R}^+$ indicates the point allocation for this sub-problem.
    \item $\mathcal{M}$ denotes the modality type, defined as:
    \begin{equation}
        \mathcal{M} = \begin{cases}
            \text{TextOnly} & \text{if } V = \emptyset \\
            \text{MultiModal} & \text{if } V \neq \emptyset
        \end{cases}
    \end{equation}
\end{itemize}

A complete olympiad problem consists of a sequence of atomic problems: $\mathcal{P}_{\text{complete}} = \{\mathcal{P}_1, \mathcal{P}_2, \ldots, \mathcal{P}_k\}$ with total score $A = \sum_{i=1}^{k} a_i$.

\subsubsection{Solution Schema}

Correspondingly, each atomic problem $\mathcal{P}_i$ has an associated solution $\mathcal{S}_i = \{C, V_{\text{ref}}, \rho\}$, where:
\begin{itemize}
    \item $C = \{c_1, c_2, \ldots, c_m\}$ denotes the set of canonical answer components extracted from official solutions, where each $c_j$ represents a distinct answer element (e.g., numerical value, chemical formula, or descriptive statement).
    \item $V_{\text{ref}} = \{v_{\text{ref}\_1}, v_{\text{ref}\_2}, \ldots, v_{\text{ref}\_l}\}$ represents the set of reference visual elements used for answer validation, such as standard molecular structures or diagrams for comparison. $V_{\text{ref}} = \emptyset$ for problems without visual answer components.
    \item $\rho: \mathcal{C}' \subseteq C \rightarrow [0, a_i]$ specifies the grading rubric that maps any subset $\mathcal{C}'$ of correct answer components to partial credit scores, with $\rho(C) = a_i$ (full credit) and $\rho(\emptyset) = 0$ (no credit).
\end{itemize}

\input{figure/overview}

\subsection{Assessment-Equivalent Reformulation (AER)}
\label{sec:aer}
The most complex aspect of benchmark construction was transforming examination questions into formats amenable to MLLM evaluation through Assessment-Equivalent Reformulation (AER). Unlike IMO or IPhO benchmarks, where reasoning-based problems naturally align with text outputs, chemistry olympiad problems often require specific modality outputs such as structural drawings, molecular diagrams, and graphical representations.

A significant challenge arose with questions requiring visual outputs: structural formula drawings, molecular annotations (atom numbering, bond labeling), reaction mechanisms (electron movement arrows), and stereochemical representations. Although advanced models (e.g., Claude, GPT-4o) can render chemical structures via HTML-based libraries or SVG generation, our validation revealed that generated outputs were difficult to verify programmatically and exhibited prohibitively high error rates due to limitations in precise spatial arrangement, stereochemical representation, and domain-specific notation.

\subsubsection{AER Components}
To address this challenge, we augment each atomic problem with three additional components through AER. The reformulated problem becomes $\mathcal{P}_{\text{AER}} = \{t, V, a, \mathcal{M}, r, \tau, \epsilon\}$, the reformulated solution becomes $\mathcal{S}_{\text{AER}} = \{C_{f}, \rho\}$, where:
\begin{itemize}
    \item $r \in \mathcal{R} = \{\text{SMILES}, \text{Numerical}, \text{Selection}, \text{Table}, \text{Text}\}$: \textbf{Requirement} specifies the expected answer format (e.g., SMILES for molecular structures, Numerical for quantitative calculations)
    \item $\tau \in \mathcal{T}_{type}$: \textbf{Problem Type} categorizes the cognitive skill being assessed (e.g., Structure Construction, Mechanism Reasoning)
    \item $\epsilon \in \mathcal{E}$: \textbf{Evaluation Method} defines how the model's output will be scored (e.g., Structure Match, Numerical Tolerance)
\end{itemize}

\vspace{-0.1cm}
\begin{tcolorbox}[
  colback=blue!3,
  colframe=blue!30,
  fonttitle=\small,
  title=\textbf{AER Example: Structure Drawing Reformulation},
  boxrule=2pt,
  arc=0.5pt
]
\small
\textbf{Original Problem ($t$)}: Draw the structure of A.

\textbf{Requirement ($r$)}: Output the SMILES string of A.

\textbf{Problem Type ($\tau$)}: Structure Construction

\textbf{Evaluation Method ($\epsilon$)}: Structure Match

\textbf{Original Answer ($C$)}: [Structural drawing image]

\textbf{Reformulated Answer ($C_{f}$)}: \texttt{CC(=O)CC}

\textbf{Grading ($\rho$)}: Binary scoring based on exact or chemically equivalent SMILES match
\vspace{-0.1cm}
\end{tcolorbox}

Through AER, we transform problems requiring visual outputs into problems requiring symbolic or textual outputs that are assessment-equivalent but computationally tractable for automated evaluation. Chemistry education experts led the reformulation process and validated the assessment equivalence. Details of the reformulation and validation procedures are provided in Appendix~\ref{app:aer}.

\subsection{Structured Visual Enhancement}
\label{sec:sve}

\textbf{Design Rationale.} Chemical visual representations, such as molecular structures, reaction schemes, and mechanism diagrams, encode complex symbolic information that requires precise interpretation. While MLLMs process these images as pixel data, there exists a hypothesis that their visual parsing capabilities may lag behind their chemical reasoning abilities. If this bottleneck exists, providing machine-readable encodings of visual content could bypass the visual interpretation step and reveal the models' true chemical problem-solving capacity.

To test this hypothesis and enable diagnostic analysis of model capabilities, we introduce an optional structured visual component $\mathcal{G}$ (``Guidance") that augments problems with symbolic encodings of visual content. The structured representations are tool-generated and manually verified for accuracy (more details in Appendix~\ref{app:structured}).

\noindent\textbf{Structured Representation.} For each visual element in a problem $P$, we construct a corresponding structured representation using specialized tools:
\begin{itemize}
    \item \textbf{Molecular structures}: Converted to SMILES or InChI strings via optical chemical structure recognition (OCSR) tools~\citep{IDEAOCSR4:online, qian2023molscribe}
    \item \textbf{Reaction schemes}: Parsed into structured reaction templates with reagents, conditions, and stoichiometry
    \item \textbf{Mechanism diagrams}: Decomposed into mechanistic steps with explicit electron flow annotations
    \item \textbf{Other visual types}: Encoded in appropriate machine-readable formats (e.g., partial SMILES for fill-in-the-blank figures, directed graphs for process flowcharts)
\end{itemize}

\noindent\textbf{Evaluation Configurations.} This yields two evaluation settings: (1) Baseline: Problems $P_{AER}$ where models receive only original text and raw images. (2) Enhanced: Problems $P_{AER} + \mathcal{G}$ where models additionally receive structured symbolic representations.

By comparing model performance across these configurations, we can isolate the contribution of visual interpretation versus chemical reasoning capabilities. This diagnostic approach helps identify whether performance limitations stem from visual understanding gaps or fundamental chemical knowledge deficits, informing potential augmentation strategies explored in Section~\ref{sec:multiagent}. The structured enhancement is provided as an optional component rather than a core benchmark requirement, allowing flexible evaluation of both pure vision-language capabilities and augmented reasoning scenarios.

\subsection{Grading Framework}

We develop a grading framework that addresses the unique challenges of chemistry assessment while maintaining compatibility with standard MLLM evaluation protocols. Our framework employs two complementary grading approaches to ensure comprehensive and reliable evaluation.

\noindent\textbf{Rubric-Based Grading.} For questions with well-defined grading rubrics in the original solutions, we systematically convert all scoring criteria into a unified deductive framework. Each question begins with full points allocated, and specific deductions are applied based on identified errors or missing components in the model response. Where applicable, we employ domain-specific validation tools such as RDKit~\citep{RDKit} for molecular structure verification, numerical tolerance checking for quantitative calculations, and pattern matching for structured outputs. The rubric-based approach ensures consistency with authentic chemistry assessment practices while enabling reproducible scoring.

\noindent\textbf{LLM-as-a-Judge Grading.} For questions where strict rubric-based grading yields binary outcomes that fail to capture partial progress, we employ LLM-as-a-Judge to provide fine-grained evaluation. For instance, when a question requires multiple correct answers and the original rubric awards zero points unless all are correct, this approach fails to distinguish between a response with one correct answer versus three correct answers. We use LLM to evaluate the semantic alignment between model responses and reference solutions, computing a normalized similarity score that reflects the degree of correctness and reasoning quality. This approach provides nuanced differentiation of model capabilities, particularly for multi-component questions where partial understanding should be reflected in the score.

\subsection{Dataset Characteristics}

The ChemO benchmark comprises 9 problems, containing 59 sub-problems across five chemical domains: organic chemistry (P1-P2), inorganic chemistry (P3, P8), physical chemistry (P4-P6), analytical chemistry (P7), and biochemistry (P9). Tab.~\ref{tab:icho_problems} provides a detailed overview of each problem's composition, reasoning requirements, and difficulty characteristics. Problems utilize diverse representations from text-only to multimodal formats (structures, spectra, graphs, mechanisms), with harder problems requiring multiple integrated representation types.

\noindent\textbf{Problem Type Distribution.} The 59 sub-problems span five reasoning categories with distinct cognitive requirements: Structure Construction (SC, 28.8\%) demands spatial and topological reasoning for molecular architecture; Quantitative Calculation (QC, 30.5\%) requires numerical computation and formula application; Qualitative Identification (QI, 25.4\%) involves pattern recognition and chemical property analysis; Tabular Enumeration (TE, 8.5\%) necessitates systematic organization of chemical information; and Mechanistic Reasoning (MR, 6.8\%) requires understanding of reaction pathways and transformation logic. This distribution reflects the diverse cognitive skills essential for advanced chemical problem-solving.

\noindent\textbf{Problem Complexity and Integration.} Problems exhibit varying levels of scope and conceptual integration. Single-focus problems (e.g., P5 with exclusively quantitative calculations) contrast with multi-faceted challenges like P8, which integrates four reasoning categories: SC$\times$2, QI$\times$4, TE$\times$1, and QC$\times$2. Difficulty levels range from Medium (P4, P7) to High (P2, P3, P6, P9), determined by factors including conceptual sophistication, multi-step reasoning depth, prerequisite knowledge requirements, and computational complexity. Higher-difficulty problems require more granular evaluation of partial progress and intermediate reasoning steps.

\subsection{Estimated Human Performance Baseline}

The IChO~\citep{IChO56:online} does not publicly disclose specific medal cutoff scores, maintaining confidentiality around performance metrics. However, our investigation of the 2021 Japan IChO theoretical exam yielded performance statistics that allow us to establish an estimated human baseline~\citep{ResultsI23:online}. The 2021 exam showed a mean score of 43.91 and a standard deviation of 24.26 (on a 100-point scale), indicating substantial spread in contestant abilities~\citep{icho2021statistics}.

Based on these statistics and standard IChO medal distributions, we can estimate the 2021 medal thresholds using z-score calculations~\citep{IChO56:online}. Assuming a normal distribution, the cutoffs are computed as:

\begin{equation}
    \text{Estimated Cutoff} = \mu + z \cdot \sigma
    \label{eq:z-score-concise}
\end{equation}

where $\mu$ is the mean, $\sigma$ is the standard deviation, and $z$ is the z-score~\citep{cheadle2003analysis} for the target percentile. For the IChO 2021 data:
\gold \textbf{Gold Medal} (top $\sim 10\%$, $z \approx 1.28$): $43.91 + 1.28 \times 24.26 \approx 75.0$ points;
\silver \textbf{Silver Medal} (top $\sim 30\%$, $z \approx 0.52$): $43.91 + 0.52 \times 24.26 \approx 56.5$ points;
\bronze \textbf{Bronze Medal} (top $\sim 60\%$, $z \approx -0.25$): $43.91 - 0.25 \times 24.26 \approx 37.8$ points.

We use this 2021-based baseline to calibrate model performance against a challenging standard, acknowledging that annual variations in exam difficulty preclude direct claims of medal achievement. Therefore, surpassing the estimated gold threshold signifies problem-solving abilities comparable to the top percentile of human contestants.

%% file: figure/overview.tex
\begin{figure*}[!htbp] 
    \centering    
    \includegraphics[width=\linewidth]{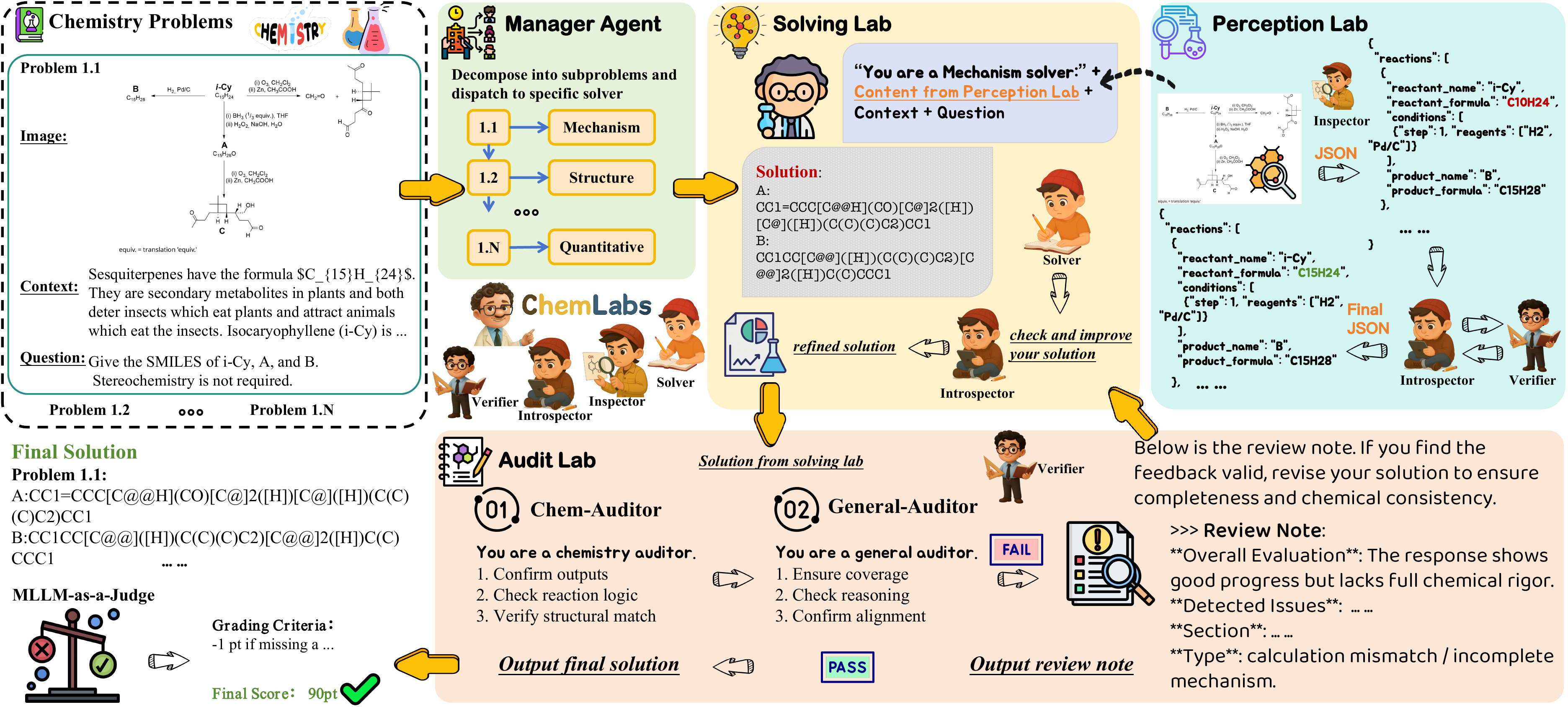}
        \caption{\textbf{Overview of \textsc{\ours}}, a hierarchical multi-agent framework for solving IChO problems. Each complete question is first received by a \textbf{manager agent}, which autonomously decomposes it into sub-tasks (e.g.,~1.1,~1.2,~1.3) and dispatches them to domain-specific solvers according to their types. Visual sub-tasks are processed through the \textbf{Perception Lab} for structured interpretation, followed by task-specific reasoning in the \textbf{Solving Lab}. The resulting answers are refined by the introspector and verified in the \textbf{Audit Lab} via \textbf{Chem-Auditor} and \textbf{General-Auditor}. This design enables adaptive task allocation, modular reasoning, and interpretable multi-agent collaboration across diverse chemical problem types.}
    \label{fig:overview}
    \vspace{-10pt}
\end{figure*} 

%% file: sec/4_agent.tex
\section{\textsc{\ours}: A Hierarchical Multi-Agent System}
\label{sec:multiagent}

\subsection{Overview of Framework}
\textsc{\ours} adopts a hierarchical multi-agent architecture inspired by how human teams collaboratively solve complex Olympiad problems, as shown in Fig.~\ref{fig:overview}. Each IChO task, often comprising several sub-questions (e.g., 1.1, 1.2, 1.3), is dispatched to the system, where a \emph{manager agent} autonomously decomposes it into sub-tasks and assigns each to the most suitable specialist. This design differs from pre-defined static workflows; instead of manually encoding routing rules, the manager leverages its reasoning ability to infer task dependencies and dynamically schedule solvers based on content, modality, and question type.

Once decomposition is complete, each sub-task enters the standard three-phase pipeline: the optional \textbf{Perception Lab} for image understanding, the \textbf{Solving Lab} for reasoning and structured solution generation, and the \textbf{Audit Lab} for verification. This framework ensures that the system can flexibly adapt to diverse chemical reasoning tasks—from structure identification to quantitative calculations—while maintaining modular and interpretable agent collaboration.

\subsection{Perception Lab}
\label{framework:perception_lab}

\noindent\textbf{Architecture.}
The Perception Lab is invoked when a sub-task requires visual interpretation without sufficient textual description. A perception agent processes the input image (reaction schemes, molecular structures, or spectroscopic data) and produces a structured textual representation that lists chemical entities, captures spatial relationships and annotations, and encodes the chemical semantics needed by the downstream solver in the Solving Lab.

\noindent\textbf{Key Contribution.}
The Perception Lab establishes a principled interface between visual chemical information and symbolic reasoning systems. By explicitly converting images into structured text rather than employing end-to-end multimodal processing, it reduces ambiguity and prevents error accumulation. This decoupling of perception and reasoning enables downstream solvers to operate on verified representations, enhancing both interpretability and reliability—particularly critical for complex IChO problems involving multiple reaction pathways or annotated spectra.

\subsection{Solving Lab}
\label{framework:solving_lab}

\noindent\textbf{Architecture.}
The Solving Lab constitutes the reasoning engine of \ours. For each sub-task from the manager agent, the dispatcher assigns one domain-specific solver from a set including Structure Construction, Quantitative Calculation, Qualitative Identification, Tabular Enumeration, and Mechanistic Reasoning. Each solver implements a specialized problem-solving strategy aligned with its domain, analogous to expert chemists employing distinct methodologies for different question types: retrosynthetic analysis for structural problems, mechanistic pathways for reaction mechanisms, and numerical derivation for quantitative tasks. Each solver first formulates an internal strategy (identifying reaction indicators, retrieving relevant equations, or analyzing spectral signatures), then generates a structured JSON solution conforming to a unified schema. The \textbf{solving-introspector} performs a single refinement pass to ensure logical consistency, format compliance, and explicit reasoning justification. The refined solution proceeds to the Audit Lab for validation.

\noindent\textbf{Key Contribution.}
In contrast to rigid pipeline architectures with predetermined module sequences, the Solving Lab enables dynamic solver invocation through autonomous task allocation. This design allows language models to exhibit adaptive reasoning, selecting solution strategies appropriate to each problem type. The unified JSON schema enforces transparency, requiring solvers to articulate their reasoning explicitly, thereby ensuring traceability and interpretability of agent behaviors.

\subsection{Audit Lab}
\label{framework:audit_lab}

\noindent\textbf{Architecture.}
The Audit Lab implements a two-stage verification protocol: the \textbf{chem-auditor} validates domain-specific correctness, while the \textbf{general-auditor} confirms overall logical integrity. The chemistry-specific stage verifies reaction stoichiometry, oxidation state consistency, mass balance, and intermediate plausibility; the general stage examines JSON format adherence, reasoning coherence, and computational accuracy. Upon detecting errors, a diagnostic report is returned to the \textbf{solving-introspector} for revision, followed by a final re-audit. Solutions proceed only after approval from both auditors.

\noindent\textbf{Key Contribution.}
The Audit Lab incorporates a dual-perspective verification framework addressing both chemical validity and logical soundness. This hierarchical approach reflects expert evaluation practices in IChO assessments—validating domain fidelity before assessing reasoning quality. Through structured error reporting and targeted refinement, the Audit Lab establishes a collaborative verification mechanism that produces reliable, interpretable, and standardized chemical reasoning outputs.

%% file: table/table_2.tex
\begin{table*}[t]
\vspace{-0.5cm}
\centering
\fontsize{8}{10}\selectfont
\setlength{\tabcolsep}{5pt}
\renewcommand{\arraystretch}{1}
\begin{tabular}{l|ccccccccc|c|c}
\toprule
\textbf{Problem} & \textbf{P1} & \textbf{P2} & \textbf{P3} & \textbf{P4} & \textbf{P5} & \textbf{P6} & \textbf{P7} & \textbf{P8} & \textbf{P9} & \textbf{Total} & \textbf{Rank} \\ 
\midrule
Original Points & 34 & 67 & 111 & 40 & 16 & 22 & 22 & 35 & 38 & 385 & \\ 
\textit{Norm.} / \textit{Sim.} & 9.0 / 1.00 & 17.0 / 1.00  & 29.0 / 1.00 & 10.0 / 1.00 & 4.0 / 1.00 & 6.0 / 1.00 & 6.0 / 1.00 & 9.0 / 1.00 & 10.0 / 1.00 & 100.0 & \\
\midrule
\multicolumn{12}{l}{\cellcolor[RGB]{230,240,255}\textit{Gemini-2.5 Pro}} \\
MLLM-Only & 6.2 / 0.69 & 15.0 / 0.86 & 13.3 / 0.46 & 9.5 / 0.91 & 4.0 / 0.99 & 5.5 / 0.96 & 2.5 / 0.41 & 7.4 / 0.82 & 7.2 / 0.72 & 70.6 & \silver \\
+ MAS & 6.8 / 0.73 & 15.2 / 0.87 & 15.8 / 0.54 & 9.7 / 0.93 & 4.0 / 0.99 & 5.6 / 0.97 & 3.2 / 0.53 & 7.7 / 0.83 & 7.4 / 0.74 & 75.4 & \gold \\
+ SVE & 7.1 / 0.76 & 15.1 / 0.87 & 19.0 / 0.66 & 9.9 / 0.95 & 4.0 / 0.99 & 5.6 / 0.95 & 4.1 / 0.68 & 7.9 / 0.87 & 7.6 / 0.78 & 80.3 & \gold \\
+ SVE \& MAS & \textbf{8.3 / 0.89} & \textbf{15.3 / 0.88} & \textbf{28.5 / 0.99} & \textbf{10.0 / 0.98} & \textbf{4.0 / 0.99} & \textbf{5.7 / 0.97} & \underline{5.3 / 0.89} & \underline{8.5 / 0.94} & 8.0 / 0.82 & \textbf{93.6} & \gold \\ 
\midrule
\multicolumn{12}{l}{\cellcolor[RGB]{230,255,240}\textit{Claude-3.7 Sonnet}} \\
MLLM-Only & 0.6 / 0.07 & 5.1 / 0.31 & 18.2 / 0.63 & 9.9 / 0.95 & 4.0 / 0.99 & 5.6 / 0.96 & 5.7 / 0.94 & 8.3 / 0.91 & 6.8 / 0.71 & 64.2 & \silver \\
+ MAS & 2.1 / 0.25 & 8.3 / 0.51 & 19.5 / 0.67 & 10.0 / 0.96 & 4.0 / 0.99 & 5.6 / 0.96 & 5.8 / 0.96 & 8.5 / 0.93 & 7.1 / 0.73 & 70.9 & \silver \\
+ SVE & 6.5 / 0.74 & 14.8 / 0.85 & 21.6 / 0.77 & 10.0 / 0.96 & 4.0 / 0.99 & 5.6 / 0.96 & 5.5 / 0.89 & 8.4 / 0.92 & 7.8 / 0.81 & 84.2 & \gold \\
+ SVE \& MAS & \underline{7.9 / 0.86} & 15.2 / 0.87 & \underline{27.4 / 0.95} & \textbf{10.0 / 0.97} & \textbf{4.0 / 0.99} & \textbf{5.7 / 0.98} & \textbf{5.7 / 0.94} & \textbf{8.9 / 0.95} & \textbf{8.4 / 0.86} & \underline{93.2} & \gold \\ 
\midrule
\multicolumn{12}{l}{\cellcolor[RGB]{255,240,230}\textit{GPT-o3}} \\
MLLM-Only & 0.3 / 0.03 & 12.7 / 0.76 & 11.8 / 0.41 & 10.0 / 0.96 & 3.8 / 0.93 & 5.6 / 0.95 & 0.8 / 0.14 & 7.1 / 0.81 & 6.9 / 0.72 & 59.0 & \silver \\
+ MAS & 1.8 / 0.22 & 13.5 / 0.79 & 13.6 / 0.49 & 10.0 / 0.97 & 3.9 / 0.95 & 5.6 / 0.96 & 1.5 / 0.27 & 7.4 / 0.84 & 7.1 / 0.74 & 64.4 & \silver \\
+ SVE & 6.6 / 0.71 & \textbf{15.3 / 0.88} & 16.7 / 0.60 & 10.0 / 0.97 & 4.0 / 0.98 & 5.6 / 0.96 & 3.3 / 0.58 & 7.7 / 0.88 & 7.4 / 0.77 & 76.6 & \gold \\
+ SVE \& MAS & 7.4 / 0.79 & \textbf{15.3 / 0.88} & 26.0 / 0.92 & \textbf{10.0 / 0.98} & \textbf{4.0 / 0.99} & \textbf{5.7 / 0.98} & 4.7 / 0.81 & 8.4 / 0.95 & \underline{7.7 / 0.80} & 89.2 & \gold \\ 
\midrule
\multicolumn{12}{l}{\cellcolor[RGB]{250,230,255}\textit{Qwen3-VL-235B-A22B-Thinking}} \\
MLLM-Only & 4.5 / 0.52 & 10.1 / 0.62 & 13.0 / 0.47 & 9.2 / 0.88 & 3.6 / 0.87 & 5.5 / 0.94 & 2.9 / 0.51 & 4.5 / 0.52 & 6.3 / 0.66 & 59.6 & \silver \\
+ MAS & 4.8 / 0.56 & 10.6 / 0.65 & 12.6 / 0.46 & 9.4 / 0.91 & 3.7 / 0.90 & 5.5 / 0.96 & 3.1 / 0.55 & 4.7 / 0.54 & 6.5 / 0.68 & 60.9 & \silver \\
+ SVE & 5.2 / 0.60 & 11.3 / 0.69 & 15.0 / 0.54 & 9.4 / 0.90 & 3.8 / 0.92 & 5.5 / 0.96 & 3.5 / 0.61 & 5.5 / 0.63 & 6.8 / 0.71 & 66.0 & \silver \\
+ SVE \& MAS & 6.5 / 0.75 & 12.5 / 0.76 & 21.6 / 0.77 & 9.6 / 0.93 & \textbf{4.0 / 0.99} & \textbf{5.7 / 0.98} & 4.5 / 0.78 & 6.8 / 0.72 & 7.1 / 0.74 & 78.3 & \gold \\
\bottomrule
\end{tabular}
\caption{Performance on the ChemO Benchmark. We evaluate four reasoning MLLMs and show that SVE combined with MAS exceeds the estimated \gold gold-medal threshold from the 2021 IChO baseline. \textit{Norm.} denotes normalized points from original scores to 100-point scale (e.g., P1: $34/385\times100\approx9.0$). \textit{Sim.} represents similarity between MLLMs output and ground truth solution as evaluated by LLM judge. SVE denotes Structured Visual Enhancement in ChemO, and MAS denotes Multi-Agent System (\ours). \textbf{Bold} indicates best result per problem. \underline{Underline} marks second best when only one method achieves the best.}
\label{tab:main}
\vspace{-10pt}
\end{table*}

%% file: sec/5_experiments.tex
\section{Experiments}
\label{sec:experiments}

\subsection{Experimental Setup}

\noindent\textbf{Tasks and Dataset.}
We evaluate MLLMs performance on the ChemO benchmark. The benchmark contains 9 problems (P1--P9) with a total of 385 points and 59 atomic sub-questions across five domains.
Each atomic problem is reformulated via Assessment-Equivalent Reformulation (AER) into $\mathcal{P}_{\text{AER}} = \{t, V, a, \mathcal{M}, r, \tau, \epsilon\}$ and $\mathcal{S}_{\text{AER}} = \{C_f, \rho\}$. The AER process ensures that problems with visual outputs are mapped to assessment-equivalent symbolic or textual outputs that can be graded automatically. 

\noindent\textbf{Models and Configurations.}
We consider four state-of-the-art MLLMs in their reasoning oriented variants:
Gemini-2.5 Pro~\citep{gemini2.5},
Claude-3.7 Sonnet~\citep{anthropic_claude3_7},
GPT-o3~\cite{openai_o3_blog},
and Qwen3-VL-235B-A22B~\citep{qwen3vl}.
All models are accessed through their official APIs and evaluated in a zero-shot setting without task-specific fine-tuning.
For each backbone we compare four configurations that progressively incorporate SVE and multi-agent system:

\begin{itemize}
    \item \textbf{MLLM-Only.}
    The baseline configuration in which the backbone model receives the AER reformulated problem $P_{\text{AER}}$ (text and images) and directly produces answers for all sub-questions with a single call. No structured visual component $\mathcal{G}$ and no multi-agent orchestration are used.

    \item \textbf{+MAS.}
    The \textsc{\ours} multi-agent system operates on the same inputs as MLLM-Only, that is, only $P_{\text{AER}}$ without structured visual enhancement. The manager agent decomposes each problem into sub-tasks, routes them to domain-specific solvers in the Solving Lab, and the Audit Lab performs chemistry-specific and general logical verification. The Perception Lab may verbalize images when needed, but no benchmark-level structured encodings $\mathcal{G}$ (such as SMILES extracted by OCSR) are provided.

    \item \textbf{+SVE.}
    This configuration adds Structured Visual Enhancement while keeping a single-agent solver. Models receive both $P_{\text{AER}}$ and the structured visual guidance $\mathcal{G}$. The backbone still operates as a single agent, but can exploit tool-generated symbolic encodings of visual content instead of relying purely on pixel-level perception.

    \item \textbf{+SVE\&MAS.}
    The full \textsc{\ours} system, where SVE is combined with the hierarchical multi-agent system.
\end{itemize}

\noindent\textbf{Evaluation Metrics.}
Tab.~\ref{tab:main} reports two metrics for each problem in the format normalized rubric-based score and LLM-as-a-Judge similarity:

\begin{itemize}
    \item \textbf{Normalized rubric-based score.}
    For each problem, the rubric-based grader returns the total points obtained under the unified deductive framework built from the official IChO rubrics. These raw points are mapped to a global 100-point scale using the original weights. The row \textit{Original Points} lists the raw allocations for P1--P9, which sum to 385. The row \textit{Norm.} shows the maximum normalized contribution of each problem on this scale
    
    \item \textbf{LLM-as-a-Judge similarity.}
    In parallel, we compute an LLM-as-a-Judge similarity score in \([0, 1]\) using an external LLM judge that evaluates the semantic alignment between the model response and the reference solution. Higher values indicate stronger agreement in both content and reasoning, and complement the rubric-based score when rubrics are coarse or binary.
\end{itemize}

The \textbf{Total} column in Tab.~\ref{tab:main} reports the aggregate rubric-based score across all 9 problems, normalized to 100 points by summing weighted contributions from P1--P9. The \textbf{Rank} column denotes relative performance within each backbone for comparison purposes and does not correspond to official IChO medals.

\subsection{Main Results on ChemO}

\noindent\textbf{Overall performance.}
As shown in Tab.~\ref{tab:main}, both Structured Visual Enhancement and the multi-agent system contribute to stronger performance, and their combination yields the best results across all four backbones. For Gemini-2.5 Pro, the total score increases from 70.6 in the MLLM-Only setting to 75.4 with MAS alone, 80.3 with SVE alone, and 93.6 with SVE\&MAS. Claude-3.7 Sonnet follows a similar trend (64.2 $\rightarrow$ 70.9 $\rightarrow$ 84.2 $\rightarrow$ 93.2), GPT-o3 improves from 59.0 to 64.4, 76.6, and 89.2, and Qwen3-VL from 59.6 to 60.9, 66.0, and 78.3. These gains indicate that ChemO is far from saturated and that both structured visual inputs and hierarchical coordination are important for solving Olympiad-level chemistry problems.

\noindent\textbf{Comparison across backbones.}
Among the four backbones, Gemini-2.5 Pro with SVE\&MAS achieves the highest ChemO score (93.6), closely followed by Claude-3.7 Sonnet (93.2), while GPT-o3 (89.2) and Qwen3-VL (78.3) lag behind but still benefit substantially from the same framework. The relative ordering is largely consistent across P1--P9, indicating that \textsc{\ours} amplifies rather than flattens differences between backbones. At the same time, weaker models such as Qwen3-VL exhibit large relative improvements over their MLLM-Only baselines, which shows that structured perception and coordinated multi-agent reasoning can significantly enhance even less capable MLLMs on challenging Olympiad-style tasks.

\noindent\textbf{Rubric-based scores, similarity, and human baseline.}
Normalized rubric-based scores and LLM-as-a-Judge similarity scores are generally aligned: configurations with higher rubric-based scores typically also have higher similarity scores, indicating that improvements reflect genuine gains in semantic correctness rather than overfitting to the rubric. The \textit{Norm.} row provides the upper bound for each problem under our normalization, and the best SVE\&MAS configurations approach these maxima on numerically focused problems such as P4 and P5. When compared to the human performance baseline derived from IChO 2021 statistics, the strongest configurations reach or exceed the estimated gold threshold on our 100-point scale. This suggests that, under our grading protocol and with access to structured guidance and multi-agent reasoning, frontier MLLMs can approach top student performance on IChO-style theoretical chemistry problems, although differences in exam year, context, and constraints mean that this comparison should be interpreted with caution.

%% file: sec/7_analysis.tex
\section{Analysis}
\label{sec:analysis}

In this section, we analyze how Structured Visual Enhancement (SVE) and the hierarchical multi-agent system \ours affect model behavior on ChemO, by comparing the four configurations MLLM-Only, +MAS, +SVE, and +SVE\&MAS in Tab.~\ref{tab:main}.

\noindent\textbf{Effect of MAS.}
The transition from MLLM-Only to +MAS isolates the effect of hierarchical orchestration in \textsc{\ours} when no benchmark-level structured visual guidance is available. Across backbones, MAS yields consistent but moderate improvements in the total score. For example, Gemini-2.5 Pro gains about 5 points (70.6 to 75.4), Claude-3.7 Sonnet gains about 6 points (64.2 to 70.9), and GPT-o3 gains about 5 points (59.0 to 64.4). Per-problem scores show that MAS particularly helps on longer problems with multiple sub-questions, such as P3 and P8, where task decomposition, specialized solvers, and auditing reduce global coherence errors and some algebraic mistakes. However, without SVE, MAS is still constrained by imperfect visual perception, so gains on structure-heavy and mechanism-heavy problems remain limited.

\noindent\textbf{Effect of SVE.}
The transition from MLLM-Only to +SVE isolates the effect of the structured visual component $\mathcal{G}$ under single-agent inference. Across all backbones, SVE improves both normalized rubric-based scores and similarity scores, particularly on visually intensive and structure-heavy problems such as P1, P3, P7, and P8. For example, Gemini-2.5 Pro on P3 improves from 13.3 to 19.0 in normalized score and from 0.46 to 0.66 in similarity, while Claude-3.7 Sonnet improves from 18.2 to 21.6 and from 0.63 to 0.75. Compared to +MAS, +SVE often yields larger absolute gains, which supports the hypothesis from Sec.~\ref{sec:sve} that visual parsing is a primary bottleneck and that explicit machine-readable encodings of diagrams allow models to leverage their stronger symbolic reasoning capabilities even without multi-agent coordination.

\noindent\textbf{Effect of MAS with SVE.}
The improvements from +SVE to +SVE\&MAS capture the contribution of the hierarchical multi-agent design in \textsc{\ours} when models also benefit from structured visual guidance. For Gemini-2.5 Pro, the total score increases from 80.3 to 93.6, with large gains on long, multi-step problems such as P3 and P7 that require coordinated reasoning over several sub-tasks. Claude-3.7 Sonnet shows a similar jump from 84.2 to 93.2, GPT-o3 from 76.6 to 89.2, and Qwen3-VL from 66.0 to 78.3. At the per-problem level, the +SVE\&MAS configuration almost always achieves the highest normalized score and similarity for a given backbone, which suggests that manager-driven decomposition, domain-specific solvers in the Solving Lab, and chemistry-aware auditing jointly reduce errors that remain even when structured visual guidance is available. The fact that weaker backbones such as Qwen3-VL benefit proportionally more from +SVE\&MAS than from +SVE alone further indicates that orchestration and verification are complementary to raw model capability on ChemO.

%% file: sec/8_conclusion.tex
\section{Conclusion}
\label{sec:conclusion}
We present ChemO, a benchmark that reformulates IChO 2025 theoretical problems into assessment-equivalent, machine-gradable tasks via AER and an optional SVE setting. Building on this benchmark, we introduce ChemLabs, a hierarchical multi-agent system that coordinates perception, domain-specific solving, and auditing for Olympiad-level chemistry. Experiments with four frontier MLLMs show that SVE and ChemLabs are complementary, with ChemLabs + SVE on Gemini-2.5 Pro achieving 93.6/100 on ChemO and surpassing an estimated human gold-medal threshold. We hope ChemO and ChemLabs provide a practical foundation for future work on rigorous evaluation and advanced multimodal reasoning in chemistry.

%% file: sec/X_suppl.tex
\clearpage
\setcounter{page}{1}
\maketitlesupplementary

\section{Data Leakage Verification}
\label{app:vert}

\subsection{Why IChO 2025?}

We deliberately build ChemO from the IChO 2025 theoretical examination rather than earlier years. Problems and solutions from IChO exams before 2025 have been widely circulated in training materials, problem compilations, and online repositories, and are therefore likely to have been included in the pretraining or finetuning data of modern large models. In contrast, the 2025 exam was released after the training cutoffs of most deployed MLLMs used in our study, which substantially reduces the risk that it appears in their training corpora.

\subsection{Leakage Testing Protocol}

To further verify that the IChO 2025 theoretical exam is not memorized by the models we evaluate, we perform a set of empirical contamination checks.

\noindent\textbf{Web surface check.}
We sample distinctive spans from the official 2025 booklet, including long problem statements and characteristic phrasings, and search them on the public web. At the time of data collection, we only observe these spans on official IChO materials, and do not find independent reposts or solution writeups that would be likely to enter training data.

\noindent\textbf{Direct recall probes.}
For each problem, we prompt models with high level cues such as  
\emph{``Consider an IChO style problem about compound i-Cy and intermediates A, B, and C. Reproduce the full problem statement.''}  
If a model had memorized the exam, it could output the exact wording or full marking scheme. In practice, models generate generic olympiad style questions rather than the true IChO 2025 text.

\noindent\textbf{Completion probes.}
We also give models partial prefixes of real problem statements and ask them to continue the text. The continuations are compared against the ground truth booklet. We do not observe long exact matches or systematic reconstruction of full questions. Numerical values, wording, and even the structure of the continuation usually deviate from the official version, which is inconsistent with direct memorization.

\noindent\textbf{Solution recall probes.}
Finally, we ask models for \emph{``publicly available solutions to the IChO 2025 theoretical exam''} and request citations. Models fail to produce the official solutions or to name concrete URLs, and instead respond with generic advice on solving IChO style problems.

\subsection{Findings}

Across these checks, we find no empirical evidence that the evaluated models have memorized the IChO 2025 theoretical exam. Combined with the fact that earlier IChO problems before 2025 are much more likely to appear in training data, this supports our choice of the 2025 exam as a low contamination source. In the remainder of this work, we therefore treat ChemO as a held out benchmark for the models we study, while acknowledging that our tests can only provide evidence against strong leakage rather than absolute guarantees.

\section{Details of Assessment-Equivalent Reformulation (AER)}
\label{app:aer}

The goal of AER is to transform original IChO problems into formats that are easy for MLLMs to answer and for us to evaluate automatically, while preserving the underlying assessment target. For each atomic problem $\mathcal{P}$, AER specifies a requirement type $r$, a problem type $\tau$, an evaluation method $\epsilon$, a canonical reformulated answer $C_{f}$, and a grading function $\rho$ that mirrors the official marking scheme.

\subsection{Structure Construction Problems}

Many IChO questions require drawing or editing chemical structures. For example, IChO 2025 Problem 1.1 asks:

\begin{ichoproblem}[IChO 2025 Problem 1.1]
\emph{Draw the structures of i-Cy, A, and B. Stereochemistry is not required.}
\end{ichoproblem}

\noindent{and Problem 1.7 asks:}

\begin{ichoproblem}[IChO 2025 Problem 1.7]
\emph{Complete the structure of intermediate X by adding double bonds in the correct places.}
\end{ichoproblem}

Directly asking MLLMs to produce line drawings or to place bonds in an image is difficult to parse and to grade programmatically. For such Structure Construction (SC) problems, we therefore choose
\begin{align*}
r        &= \text{SMILES},\\
\tau     &= \text{Structure Construction},\\
\epsilon &= \text{Structure Match}.
\end{align*}

Chemistry experts first redraw the target structures in ChemDraw\footnote{ChemDraw, \url{https://revvitysignals.com/products/research/chemdraw}.} and export SMILES strings. These SMILES are canonicalized with a cheminformatics toolkit and stored as $C_f$ for each species (for example, i-Cy, A, B, X). During evaluation, the model is prompted to:

\begin{itemize}
    \item Output the SMILES of i-Cy.
    \item Output the SMILES of intermediate X after adding all required double bonds.
\end{itemize}

The grading function $\rho$ parses the predicted SMILES, reconstructs the molecule, and checks graph isomorphism against the reference structure. When the original problem ignores stereochemistry, evaluation also ignores stereochemical labels. For multi part questions such as Problem 1.1, we aggregate the scores for i-Cy, A, and B according to the original point allocation.

\subsection{Stereocentre Assignment Problems}

Some questions require both visual identification of stereocentres and assignment of R/S descriptors. IChO 2025 Problem 1.2 states:

\begin{ichoproblem}[IChO 2025 Problem 1.2]
\emph{Circle the stereocentres in compound C and assign them as R or S.}
\end{ichoproblem}

For models, directly drawing circles on the image and then evaluating their positions is highly unreliable and difficult to score. We therefore convert the structure of C to a canonical SMILES string and reformulate the task as a textual reasoning problem over a numbered carbon skeleton. The question side becomes:

\begin{itemize}
    \item \textbf{Context:} Circle the stereocentres in compound C and assign them as R or S.
    \item \textbf{Image:} the original structure of C.
    \item \textbf{Parsing note:} the structure has been converted to a SMILES string.
    \item \textbf{SMILES:} canonical SMILES of C.
    \item \textbf{Requirement:}
    \begin{enumerate}
        \item Count only carbon atoms (C) in the SMILES string from left to right, ignoring all other atoms such as H, O, N. Label them sequentially as C-1, C-2, C-3, and so on.
        \item Assign R or S configuration to each stereocentre.
        \item Output in the format \texttt{C-X: R/S}, separated by commas.
    \end{enumerate}
\end{itemize}

The canonical answer $C_f$ is then the list of stereocentres and their descriptors:
\[
C_f = \text{\texttt{C-1: R, C-6: S, C-7: R}}.
\]

This reformulation preserves the original assessment target, since the model must still locate all stereocentres and assign the correct R/S descriptors, but the answer is expressed in a compact textual format that is easy to parse and to grade automatically.

\paragraph{Summary.}
In all cases, AER removes the need for models to produce free-form drawings or complex visual annotations, while preserving the original constructs tested by the olympiad problems and enabling scalable, automated evaluation.

\section{Example of Structured Visual Enhancement (SVE)}
\label{app:structured}

To illustrate the structured visual component $\mathcal{G}$ introduced in Section~\ref{sec:sve}, we provide a concrete example taken from Problem~1 of the IChO~2025 theoretical exam. Chemistry experts manually redrew all molecules in ChemDraw based on the original reaction scheme and exported the corresponding SMILES strings, which were then used to construct the structured representation. The example links the original reaction scheme image to its corresponding machine readable representation. The image shows the interconversion of \textit{i}-Cy, A, B, and C under different conditions, and the JSON encodes reactant and product identities, overall formulas, and stepwise reagents for each transformation.

\begin{figure}[h]
    \centering
    \includegraphics[width=0.8\linewidth]{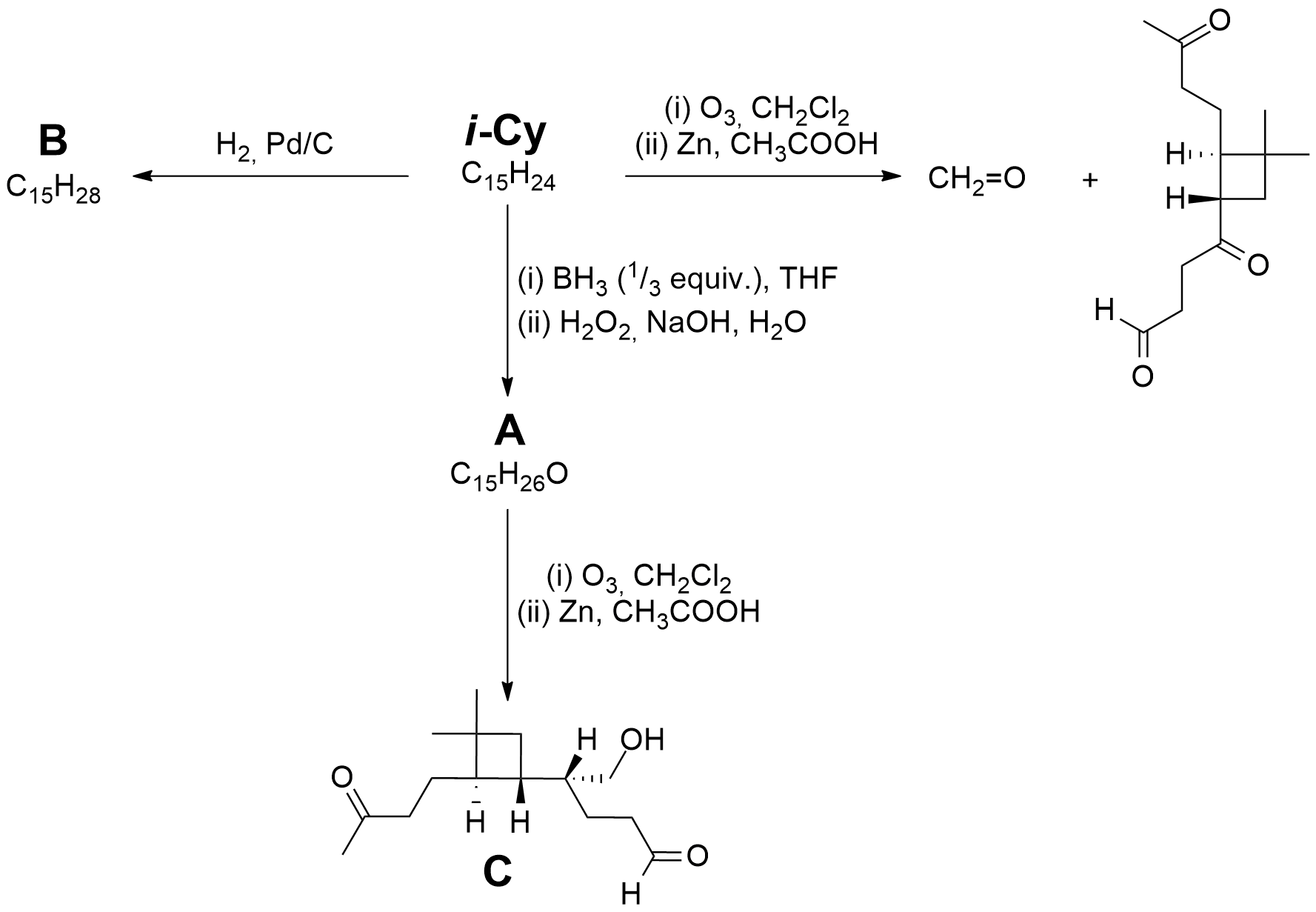}
    \caption{Example reaction scheme used to construct the structured visual guidance $\mathcal{G}$. The scheme, taken from Problem~1 of the IChO~2025 theoretical exam, depicts the transformations of \textit{i}-Cy into B, A, and C under different reaction conditions.}
    \label{fig:structured-example}
\end{figure}

\begin{chemjson}
{
  "reactions": [
    {
      "reactant_name": "i-Cy",
      "reactant_formula": "C15H24",
      "conditions": [
        {"step": 1, "reagents": ["H2", "Pd/C"]}
      ],
      "product_name": "B",
      "product_formula": "C15H28"
    },
    {
      "reactant_name": "i-Cy",
      "reactant_formula": "C15H24",
      "conditions": [
        {"step": 1, "reagents": ["BH3 (1/3 equiv.)", "THF"]},
        {"step": 2, "reagents": ["H2O2", "NaOH", "H2O"]}
      ],
      "product_name": "A",
      "product_formula": "C15H26O"
    },
    {
      "reactant_name": "i-Cy",
      "reactant_formula": "C15H24",
      "conditions": [
        {"step": 1, "reagents": ["O3", "CH2Cl2"]},
        {"step": 2, "reagents": ["Zn", "CH3COOH"]}
      ],
      "product_smiles": [
        "C=O",
        "CC1(C)[C@]([C@@]([H])(C(CCC([H])=O)=O)C1)([H])CCC(C)=O"
      ]
    },
    {
      "reactant_name": "A",
      "reactant_formula": "C15H26O",
      "conditions": [
        {"step": 1, "reagents": ["O3", "CH2Cl2"]},
        {"step": 2, "reagents": ["Zn", "CH3COOH"]}
      ],
      "product_name": "C",
      "product_smiles": "CC1(C)[C@]([C@@]([H])([C@](CO)(CCC([H])=O)[H])C1)([H])CCC(C)=O"
    }
  ]
}
\end{chemjson}